\documentclass[10pt, a4paper]{article}

\usepackage[final]{lrec2026} 
\usepackage{comment}
\usepackage{graphicx}
\usepackage{multirow}
\usepackage{makecell}
\usepackage[table]{xcolor}
\usepackage{tabularx}
\usepackage{amsmath}
\usepackage{array}
\usepackage{subcaption}

\usepackage[T1]{fontenc}
\usepackage[utf8]{inputenc}

\newcolumntype{P}[1]{>{\raggedright\arraybackslash}p{#1}} 
\newcolumntype{C}[1]{>{\centering\arraybackslash}p{#1}}   

\usepackage{listings}
\usepackage{hyperref}

\usepackage[most]{tcolorbox} 

\newcounter{doccnt} 
\newtcolorbox{DocBox}{ 
  colback=gray!10!white,
  colframe=gray!50!white,
	left=0mm,
	right=0mm,top=0mm,bottom=0mm, size=small,
  title={\sffamily\bfseries \small \hfill {Example Document \arabic{doccnt}}}
}

\definecolor{FP}{RGB}{255, 69, 0} 
\definecolor{onto_id}{RGB}{95, 112, 227} 
\definecolor{relevant}{RGB}{255, 215, 0}
\definecolor{correct}{RGB}{50, 205, 50}

\title{An Extreme Multi-label Text Classification (XMTC) Library Dataset: What if we took ``Use of Practical AI in Digital Libraries'' seriously?}

\name{Jennifer D'Souza$^{1}$, Sameer Sadruddin$^{1}$, Maximilian Kähler$^{2}$, Andrea Salfinger$^{3}$, \\
{\bf \large Luca Zaccagna$^{3}$, Francesca Incitti$^{3}$, Lauro Snidaro$^{3}$, Osma Suominen$^{4}$}}

\address{%
$^{1}$TIB Leibniz Information Centre for Science and Technology, Germany\\
$^{2}$Deutsche Nationalbibliothek, Germany\\
$^{3}$University of Udine, Italy\\
$^{4}$National Library of Finland, Finland\\
\{jennifer.dsouza,sameer.sadruddin\}@tib.eu
}

\abstract{
Subject indexing is vital for discovery but hard to sustain at scale and across languages. We release a large bilingual (English/German) corpus of catalog records annotated with the Integrated Authority File (GND), plus a machine-actionable GND taxonomy. The resource enables ontology-aware multi-label classification, mapping text to authority terms, and agent-assisted cataloging with reproducible, authority-grounded evaluation. We provide a brief statistical profile and qualitative error analyses of three systems. We invite the community to assess not only accuracy but usefulness and transparency, toward authority-anchored AI co-pilots that amplify catalogers’ work.
 \\ \newline 
}

\begin{document}

\maketitleabstract

\section{Introduction}

Libraries have long relied on expert subject indexing to make collections findable, interoperable, and durable. Yet the rapidly growing, multilingual volume of library catalog records increasingly strains purely manual indexing workflows. At the same time, large language models (LLMs) and emerging agentic pipelines promise support—but they must be grounded in authoritative vocabularies, auditable, and evaluated in library terms rather than by generic text-classification scores. We present a machine-learning-ready resource that directly addresses this gap: a bilingual (English/German), multi-domain corpus of catalog records indexed with subjects from the German Integrated Authority File (Gemeinsame Normdatei, GND), released together with a machine-actionable version of the GND subject taxonomy and predefined train/dev/test splits. The goal is not merely scale, but structured scale—where every prediction links to a controlled vocabulary that libraries already trust.

This resource is designed to help the community interrogate practical questions that matter for library science in the LLM era: \textit{How should automated systems align free text to controlled vocabularies while preserving provenance and authority control? What counts as “useful” assistance—top-k quality at the point of description, hierarchical coherence, explainable rationales, or cataloger effort saved? How can models cope with long-tail subjects, multilingual variation, and distribution shift across domains and time? Where do agents best fit in human-in-the-loop workflows (triage, suggestion, validation)}, and \textit{how should we measure their impact beyond batch metrics?} By anchoring experiments in an operational taxonomy, the dataset enables studies of vocabulary grounding, cross-lingual consistency, polysemy and variant labels, and reliability under realistic label sparsity—questions that generic XMTC benchmarks only partially surface.

At a high level, our contribution pairs real catalog records with stable links to authoritative subject concepts and packages them for reproducible evaluation. This enables ontology-aware multi-label classification, retrieval-augmented mapping from free text to authority terms, and agent workflows that combine retrieval, suggestion, and curator feedback—evaluated with protocols that reflect cataloging realities (e.g., usefulness and hierarchical consistency at the top of the record). We outline the resource, its construction and splits, and initial analyses and baselines, and we position the paper as a \textbf{guidebook} to the dataset: we statistically explore it to surface considerations for framing machine-learning solutions, and we conclude with qualitative error analyses of three systems developed on our data—inviting the LREC community to test, compare, and \textit{reflect} on what successful, trustworthy AI assistance for subject indexing should look like. This resource is released as \textbf{TIB-SID} (TIB Subject Indexing Dataset), a bilingual (English/German), multi-domain corpus of 136k catalog records annotated with GND subjects, available at \url{https://github.com/sciknoworg/tib-sid} (CC BY 4.0 license).

The rest of the paper is structured as follows: \autoref{sec:rel-work} reviews related work, \autoref{sec:our-dataset} describes our dataset, and \autoref{sec:systems} presents three systems trained on it along with a brief qualitative analysis. The article concludes in \autoref{sec:conclude}.

\section{Related Work}
\label{sec:rel-work}

\textbf{Extreme Multi-label Text Classification (XMTC) Datasets.} XMTC benchmarks \cite{Bhatia16} involve assigning items to very large label spaces with highly skewed long-tail distributions \cite{zhang2023long}. Notable examples include \textit{Wiki-500K} (1.8M Wikipedia entries, 500K categories) and \textit{AmazonCat-13K} (~1.5M product descriptions, 13{,}330 categories). In both, most labels are extremely rare (e.g., only ~2\% of Wiki-500K labels occur more than 100 times; ~30\% of AmazonCat labels appear in fewer than 10 samples) \cite{yu2022pecos,zhang2023long}. Smaller domain-specific corpora such as \textit{EurLex-4K} (19{,}000 EU legal documents, ~4{,}000 EuroVoc subjects) show similar long-tail behavior. These datasets exemplify the statistical challenges of large label spaces but typically rely on generic or user-defined categories. In contrast, our dataset represents fine-grained, technically specialized subject annotations on library records over a wide range of research domains.

\textbf{Biomedical Indexing with Large Taxonomies.} The biomedical domain has pioneered large-scale semantic indexing with corpora such as \textit{BioASQ} \cite{tsatsaronis2015overview,krithara2023bioasq}, which provides millions of PubMed abstracts annotated with \textit{Medical Subject Headings} (MeSH), a hierarchical vocabulary of about 30{,}000 descriptors \cite{national2000medical}. Each article is typically assigned 12–13 MeSH terms, framing the task as an XMTC problem. The multilingual \textit{MESINESP} corpus extends this approach to Spanish and Portuguese using DeCS \cite{gasco2021overview-mesinesp}. Together, these benchmarks demonstrate how controlled domain taxonomies can support large-scale automated subject indexing \cite{bioask-tenyears}.

\textbf{Library Cataloging and Multilingual Linked Data.} Multilingual subject-indexing datasets are only partially addressed in prior work, and openly ML-ready benchmarks are scarce. Notable efforts include the \href{https://portal.ehri-project.eu/}{EHRI} Multilingual Subject Indexing Test Dataset—Holocaust archival descriptions labeled with a 900-term vocabulary in 12 languages \cite{ehri,dermentzi-ehri-hal-05142136}; FAO’s \href{https://agris.fao.org/}{AGRIS} (16M+ records, 123 languages) indexed with the multilingual \textit{AGROVOC} thesaurus \cite{caracciolo2013agrovoc,panoutsopoulos2022data}; and Europeana, the pan-European cultural-heritage aggregator, enriches records with multilingual subject links to controlled vocabularies such as the Virtual International Authority File (VIAF), EuroVoc, or the Art \& Architecture Thesaurus (AAT), available via data dumps and SPARQL endpoints \cite{isaac2013europeana,stiller2016multilingual}. National initiatives—the British Library’s British National Bibliography (BNB) as Linked Open Data \cite{deliot2014publishing} and the Bibliothèque nationale de France’s (BnF) \href{https://data.bnf.fr/en/vocabulary}{Répertoire d’autorité-matière encyclopédique et alphabétique unifié (RAMEAU)} vocabulary—likewise expose large SKOS/RDF authority datasets. However, these resources typically lack standardized train/test splits and evaluation metrics. 

\textbf{Comparative Outlook – Our Dataset.} Unlike prior XMTC benchmarks with flat or user-defined labels and library linked-data releases without ML-ready splits, our corpus pairs XMTC-scale long tails with the structured GND taxonomy, bridging large-scale text classification and knowledge organization. It provides a bilingual, ML-ready benchmark with predefined train/dev/test splits and stable GND links for ontology-aware modeling, multilingual evaluation, and reproducible comparisons.

\begin{figure*}[!htb]
\centering

\begin{minipage}[t]{.49\textwidth}
\begin{lstlisting}
{
 "Code": "gnd:4381803-1",
 "Classification Number": "19.3",
 "Classification Name": "Hydrologie, Meereskunde",
 "Name": "TOC",
 "Alternate Name": ["Total organic carbon","Gesamter Organischer Kohlenstoff","Abwasseranalyse,Kennzahl"],
 "Related Subjects": ["Summenparameter"],
 "Source": "B 1986"
}
\end{lstlisting}
\end{minipage}\hfill%
\begin{minipage}[t]{.49\textwidth}
\begin{lstlisting}
{
 "Code": "gnd:4139622-4",
 "Classification Number": "27.7",
 "Classification Name": "Allgemeine Therapie",
 "Name": "Naturheilverfahren",
 "Alternate Name": ["Biologische Heilweise","Naturheilweise"],
 "Related Subjects": ["Erfahrungsheilkunde","Naturheilkunde"],
 "Source": "Reallex. Med."
}
\end{lstlisting}
\end{minipage}

\vspace{0.35em}

\begin{minipage}[t]{.34\textwidth}
\begin{lstlisting}
{
 "Code": "gnd:4146660-3",
 "Classification Number": "21.4",
 "Classification Name": "Elementarteilchen, Kern-, Atom-, Molekularphysik",
 "Name": "Brom-75",
 "Alternate Name": ["Brom 75"],
 "Related Subjects": ["Bromisotop"],
 "Source": "Römpp (9.Aufl.)"
}
\end{lstlisting}
\end{minipage}\hfill%
\begin{minipage}[t]{.64\textwidth}
\begin{lstlisting}
{
 "Code": "gnd:7576879-3",
 "Classification Number": "18",
 "Classification Name": "Natur, Naturwissenschaften allgemein",
 "Name": "Copley-Medaille",
 "Alternate Name": ["Copleymedaille","Copley Medal"],
 "Related Subjects": ["Naturwissenschaften","Preis,Auszeichnung"],
 "Source": "Wikipedia",
 "Definition": "Seit 1731 jährlich von der Royal Society in London vergeben, benannt nach Godfrey Copley."
}
\end{lstlisting}
\end{minipage}

\vspace{-0.25em}
\caption{Four example GND records in our internal JSON representation.}
\label{fig:gnd-json-examples}
\end{figure*}

\section{Our Subject Indexing Dataset}
\label{sec:our-dataset}

This section introduces our subject indexing dataset, which consists of two parts: the taxonomy (\autoref{sec:taxonomy}) and the library records (\autoref{sec:library-records}) that are subject-indexed using it.

\subsection{The Subject Indexing Taxonomy}
\label{sec:taxonomy}

As our subject indexing taxonomy, we use the \href{https://data.dnb.de/GND/}{GND} (Gemeinsame Normdatei / Integrated Authority File), a set of integrated authority files created by the \href{https://www.dnb.de/EN/Home/home_node.html}{German National Library} and widely adopted by German-speaking libraries to catalog and link entities such as people, organizations, topics, and works. Among these, we specifically rely on the \textit{Sachbegriff} (subject terms).

Readers can obtain the GND Sachbegriff by following our how-to guide\footnote{\href{https://github.com/sciknoworg/tib-sid/tree/main/GND/how-to}{tib-sid/GND/how-to}}, which explains how to download the latest \texttt{authorities-gnd-sachbegriff\_dnbmarc.mrc.xml.gz} file. It is encoded in MARC~21—an international metadata standard using numeric tags (e.g., 021 for identifier, 550 for related subjects). While highly interoperable across library systems, MARC~21 is not human-readable, so we converted it into a structured JSON format\footnote{\href{https://github.com/sciknoworg/tib-sid/blob/main/GND/subjects-taxonomy/GND-subjects.json}{tib-sid/GND/subjects-taxonomy/GND-subjects.json}} using our \href{https://github.com/sciknoworg/tib-sid/blob/main/GND/subjects-taxonomy/schema.json}{internal schema}. The resulting file contains 207{,}001 unique subjects, each represented as a standardized JSON object with fields mapped from MARC~21, including the GND identifier (\texttt{Code}), classification number and name, preferred term (\texttt{Name}), variant labels (\texttt{Alternate Name}), related subjects, and source, with optional entries for \texttt{Definition} and \texttt{Source URL}. When considering \href{https://github.com/sciknoworg/tib-sid/blob/main/GND/subjects-taxonomy/README.md}{property coverage}, all records include the core fields, while contextual information varies—\texttt{Alternate Name} appears in about half, \texttt{Related Subjects} in 80\%, and \texttt{Definition} in 27\%. This matters because richer contextual information within the taxonomy improves subject disambiguation, and thus enhances downstream subject indexing performance.

\autoref{fig:gnd-json-examples} illustrates four example subject terms from our JSON file: \textit{TOC} (total organic carbon) in hydrology, \textit{Naturheilverfahren} (naturopathy) in medicine, \textit{Brom-75} (bromine isotope) in physics, and \textit{Copley-Medaille} (Copley Medal) in the natural sciences. 
Each record captures both the preferred term and its variants (e.g., \textit{TOC} lists “Total organic carbon,” and \textit{Naturheilverfahren} [naturopathy] lists “Biologische Heilweise” [biological healing method] and “Naturheilweise” [natural healing method]) along with related subjects such as \textit{Summenparameter} [aggregate parameter] or \textit{Naturheilkunde} [natural medicine].
The \texttt{Source} field cites the reference from which the term originates (e.g., \textit{B~1986}, \textit{Reallex.~Med.}, or \textit{Wikipedia}), while some records also include a short explanatory \texttt{Definition}, such as the description of the Copley Medal as an annual scientific award by the Royal Society of London. Thus the structured schema preserves both lexical and contextual information.

We also provide a script\footnote{\href{https://github.com/sciknoworg/tib-sid/blob/main/GND/scripts/convert_to_skos.py}{tib-sid/GND/scripts/convert\_to\_skos.py}} to export the GND taxonomy to \href{https://www.w3.org/TR/skos-reference/}{SKOS (Simple Knowledge Organization System) format}, a W3C standard for representing controlled vocabularies in RDF. 

Next, we introduce our library records dataset annotated based on the GND. While the records appear in German or English, the GND subject terms themselves are predominantly in German, with English alternate names occasionally listed.

\subsection{Our Library Records Dataset}
\label{sec:library-records}

Our library-record corpus is derived from the open-data collection of the \href{https://www.tib.eu/en/tib/profile}{TIB -- Leibniz Information Centre for Science and Technology}. At the time of dataset construction, the TIB catalog comprised $\sim$5.7M bibliographic records and continued to grow, of which an open dump of $\sim$200{,}000 records from the TIB bibliographic holdings (TIBKAT data) was periodically released via the \href{https://www.tib.eu/en/services/open-data}{TIB open-data portal} (listed under \emph{``The metadata as a dump''} as \emph{TIBKAT data and metadata of freely available electronic collections}). The dump includes metadata from freely available electronic collections and the \href{https://av.tib.eu/}{TIB AV-Portal} (including thumbnails), and is released under CC0~1.0, allowing unrestricted reuse. To obtain a machine-learning-ready dataset from this heterogeneous and partially sparse raw dump, we applied four preprocessing steps: (i) language identification using \texttt{langdetect}\footnote{\url{https://pypi.org/project/langdetect/}}, which detected 48 languages in the raw dump (top five: German 108{,}637; English 76{,}735; French 1{,}741; Indonesian 945; Spanish 311)\footnote{Language tags in the raw dump are frequently missing or noisy; we therefore used automatic language detection. Some records remain mixed-language, but this limited noise reflects real-world metadata and does not hinder system development.}, followed by retaining only the two predominant languages, German and English; (ii) removal of records without abstracts; (iii) pruning of infrequent and/or unsuitable record types (e.g., periodicals, chapters); and (iv) removal of records without GND subject annotations, operationalized as missing the \texttt{dcterms:subject} tag.

\begin{table}[!htb]
\centering
\setlength{\tabcolsep}{2.5pt}
\resizebox{\columnwidth}{!}{%
\begin{tabular}{lrrr}
\hline
\textbf{Type/Split} & \textbf{en} & \textbf{de} & \textbf{Total} \\ \hline

\textbf{Article} & 1,738 (1.27) & 10 (0.01) & 1,748 (1.28) \\
\hspace{3mm}train & 1,103 (0.81) & 7 (0.01) & 1,110 (0.81) \\
\hspace{3mm}dev & 212 (0.16) & 2 (0.00) & 214 (0.16) \\
\hspace{3mm}test & 423 (0.31) & 1 (0.00) & 424 (0.31) \\

\rowcolor{blue!5}\textbf{Book} & \cellcolor{blue!20}44,877 (32.86) & \cellcolor{blue!30}55,704 (40.79) & \cellcolor{blue!45}100,581 (73.65) \\
\hspace{3mm}\cellcolor{cyan!10}train & \cellcolor{cyan!20}30,545 (22.37) & \cellcolor{cyan!30}35,899 (26.29) & \cellcolor{cyan!40}66,444 (48.65) \\
\hspace{3mm}dev & 7,143 (5.23) & 7,466 (5.47) & 14,609 (10.70) \\
\hspace{3mm}test & 7,189 (5.26) & 12,339 (9.03) & 19,528 (14.30) \\

\textbf{Conference} & 6,032 (4.42) & 3,693 (2.70) & 9,725 (7.12) \\
\hspace{3mm}train & 4,237 (3.10) & 2,339 (1.71) & 6,576 (4.82) \\
\hspace{3mm}dev & 1,015 (0.74) & 484 (0.35) & 1,499 (1.10) \\
\hspace{3mm}test & 780 (0.57) & 870 (0.64) & 1,650 (1.21) \\

\textbf{Report} & 2,131 (1.56) & 2,509 (1.84) & 4,640 (3.40) \\
\hspace{3mm}train & 1,463 (1.07) & 1,653 (1.21) & 3,116 (2.28) \\
\hspace{3mm}dev & 348 (0.25) & 372 (0.27) & 720 (0.53) \\
\hspace{3mm}test & 320 (0.23) & 484 (0.35) & 804 (0.59) \\

\textbf{Thesis} & 5,764 (4.22) & 14,111 (10.33) & 19,875 (14.55) \\
\hspace{3mm}train & 3,985 (2.92) & 9,221 (6.75) & 13,206 (9.67) \\
\hspace{3mm}dev & 954 (0.70) & 1,953 (1.43) & 2,907 (2.13) \\
\hspace{3mm}test & 825 (0.60) & 2,937 (2.15) & 3,762 (2.75) \\ \hline

\rowcolor{gray!10}\textbf{Total} & \cellcolor{gray!30}60,542 (44.34) & \cellcolor{gray!40}76,027 (55.66) & \cellcolor{gray!70}136,569 (100.00) \\ \hline
\end{tabular}
}
\caption{Record counts and relative shares (\%) by split, type, and language for the updated dataset. Blue: type-level shares; Cyan: split-level (Total column); Gray: totals. Cells $\geq$ 15\% are shaded.}
\label{tab:record_counts}
\end{table}

Resultingly, the cleaned collection contains 136{,}569 records, placing it among the largest bilingual, multi-domain XMTC corpora of cataloged library records. \href{https://github.com/sciknoworg/tib-sid/blob/main/library-records-dataset/data-statistics/yearwise_record_counts_by_split.csv}{Binned by decade}, the corpus is overwhelmingly modern: $\sim$0.8k (1970s) $\rightarrow$ 3.2k (1980s) $\rightarrow$ 8.6k (1990s) $\rightarrow$ 33.2k (2000s), peaking at 65.8k (2010s) with $\sim$24.6k so far in the 2020s (2024 partial), indicating rapid mid-2010s growth and a recent taper.

\autoref{tab:record_counts} summarizes our released library catalog dataset by record count and percentage of the full collection. It is organized by the five main types of records cataloged at this library, viz. Article (1,748), Book (100,581), Conference (9,725), Report (4,640), and Thesis (19,875). Another level of organization is the language of the records, viz. English (en; 60,542) and German (de; 76,027). To support ML research on subject indexing, we release the \href{https://github.com/sciknoworg/tib-sid/tree/main/library-records-dataset}{dataset} with predefined train/dev/test splits balanced by record type and language (90{,}452 / 19{,}949 / 26{,}168 records).

\begin{table*}[!htb]
\centering
\small
\setlength{\tabcolsep}{3pt}%
\renewcommand{\arraystretch}{1.08}%
\resizebox{\textwidth}{!}{%
\begin{tabular}{|p{6.0cm}|c|c|p{3cm}|p{6.5cm}|}
\hline
\makecell{\textbf{Record (title)}\\[-1pt]\footnotesize(Linked)} &
\textbf{Lang} & \textbf{Type} & \textbf{Domains} & \textbf{Subjects} \\ \hline

\href{https://github.com/sciknoworg/tib-sid/blob/main/library-records-dataset/data/dev/Article/en/3A1831632667.jsonld}{\textit{Chapter 86: Explaining the Comparative Statics in Step-Level Public Good Games}} &
en & Article & Economics & Experiment; Wirtschaftswissenschaften; Wirtschaftsforschung; Methodologie; Experimentelle Wirtschaftsforschung \\ \hline

\href{https://github.com/sciknoworg/tib-sid/blob/main/library-records-dataset/data/dev/Book/de/3A010234136.jsonld}{\textit{Das Beil von Wandsbek: Roman; 1938--1943}} &
de & Book &
Literature Studies &
Nationalsozialismus \\ \hline

\href{https://github.com/sciknoworg/tib-sid/blob/main/library-records-dataset/data/dev/Book/en/3A086058096.jsonld}{\textit{Window on Freedom: Race, Civil Rights, and Foreign Affairs, 1945--1988}} &
en & Book & Social Sciences; History & Au\ss enpolitik; Rassendiskriminierung; Menschenrecht; Schwarze \\ \hline

\href{https://github.com/sciknoworg/tib-sid/blob/main/library-records-dataset/data/dev/Conference/de/3A1005166315.jsonld}{\textit{Sicherung des Familieneinflusses in Familienunternehmen: Symposium \dots\ (6./7.\ Okt.\ 2016)}} &
de & Conf. &
Law & Einfluss; Familienbetrieb; Familie \\ \hline

\href{https://github.com/sciknoworg/tib-sid/blob/main/library-records-dataset/data/dev/Thesis/en/3A1011059754.jsonld}{\textit{Charge Carrier Recombination and Open Circuit Voltage in Organic Solar Cells: From Bilayer Model Systems to Hybrid Multi-junctions}} &
en & Thesis & Electrical engineering; Physics & Organische Solarzelle; Reihenschaltung; Mehrschichtsystem; Fluor; Donator (Chemie); Dotierung; Rekombination \\ \hline
\end{tabular}%
}
\caption{Representative examples of multi-domain, multi-subject annotations from the library records dataset. Only the \texttt{title} metadata are shown for compactness; links point to the full JSON-LD records.}
\label{tab:dataset_examples}
\end{table*}

Each library record in our dataset is represented in \texttt{json-ld}, the JSON serialization of Linked Data, which encodes bibliographic metadata as machine-readable triples. Most records include both a title and abstract, providing crucial input for machine-learning systems for subject indexing. Across record types, \href{https://github.com/sciknoworg/tib-sid/blob/main/library-records-dataset/data-statistics/abstract_stats.pdf}{abstract lengths} average 100--150 tokens; English abstracts are generally longer than German. Theses are longest in both languages (often $>140$ tokens), while Reports are shortest ($\sim$90--110). Other core bibliographic properties such as creator (\texttt{dcterms:creator}), contributor (\texttt{dcterms:contributor}), publisher (\texttt{dc:publisher}), issuing institution and place of publication (\texttt{rda:P60163}), and date of issue (\texttt{dcterms:issued}) are expressed as persistent URIs using standard vocabularies including \texttt{Dublin Core}, \texttt{BIBO}, and \texttt{RDA}. For each record, annotated GND subject tags are recorded under \texttt{dcterms:subject}, linking to controlled concept identifiers in the GND authority file. E.g., the record \href{https://github.com/sciknoworg/tib-sid/blob/main/library-records-dataset/data/dev/Conference/en/3A019447183.jsonld}{\texttt{dev/Conference/en/3A019447183.jsonld}} describes a conference paper authored by \emph{Tim Bedford} (\texttt{dcterms:creator $\rightarrow$ gnd:171970268}), published by \emph{Oxford University Press} (\texttt{dc:publisher}), and linked to subjects such as \emph{Ergodentheorie} (\texttt{dcterms:subject $\rightarrow$ gnd:4015246-7}) and \emph{Hyperbolische Geometrie} (\texttt{gnd:4161041-6}). Similarly, the record \href{https://github.com/sciknoworg/tib-sid/blob/main/library-records-dataset/data/dev/Thesis/de/3A011101717.jsonld}{\texttt{dev/Thesis/de/3A011101717.jsonld}} represents a doctoral dissertation by \emph{Reinhard O.~Greiling} (\texttt{dcterms:creator $\rightarrow$ gnd:1090799454}), published by \emph{Lang Verlag} (\texttt{dc:publisher}) in \emph{Frankfurt am Main} (\texttt{rda:P60163}), with subject links to \emph{Schwarzschiefer} (\texttt{dcterms:subject $\rightarrow$ gnd:4126782-5}) and \emph{Silur} (\texttt{gnd:4181434-4}). This representation supports interoperable bibliographic and semantic processing across linked-data resources.

\noindent\textbf{How are the records cataloged?} A team of 17 subject specialists at the TIB covers a \href{https://github.com/sciknoworg/tib-sid/blob/main/28_domains_list.csv}{predefined set of 28 domains} and assigns subject annotations to records in the TIB national library catalog, ensuring broad and expert-curated coverage. Domain assignment is usually semi-automatic: new records often arrive with domain metadata through the national library network; otherwise, an in-house \href{https://annif.org/}{ANNIF} instance is used \cite{suominen2019annif,suominen_2023_8262313}. Records may carry multiple domains (\href{https://github.com/sciknoworg/tib-sid/blob/main/library-records-dataset/data-statistics/domain_annotation_frequencies.csv}{range 1--7; mean 1.5}, with 7 as an outlier). The \href{https://github.com/sciknoworg/tib-sid/blob/main/library-records-dataset/data-statistics/domain_occurrences_by_split_and_domain.csv}{distribution is top-heavy}: Social Sciences ($\sim$24k), Economics ($\sim$24k), and Educational Science ($\sim$18k) dominate, alongside solid STEM representation (e.g., Computer Science $\sim$13k; Mathematics $\sim$12k) and a notable “Other” bucket ($\sim$5k). Specialized areas form a long tail (e.g., Mining 166; Medical Technology 589; Sports Science $\sim$1.3k; Materials Science $\sim$1.7k), indicating domain imbalance.

In practice, libraries index content with controlled vocabularies; in Germany, the GND is used for subject cataloging. Subject librarians assign GND terms from titles, abstracts, and—where available—full text, in a cooperative workflow across institutions. This work is done in a cooperative process in various libraries and in different national library networks. Given sustained growth ($\approx$15k newly indexed titles/month), this work is substantial; NLP/AI offers clear potential to support it. In our dataset, records have on average three GND subjects; \href{https://github.com/sciknoworg/tib-sid/blob/main/library-records-dataset/data-statistics/subject_annotation_frequencies.csv}{the range is 1–39} (the upper extreme is rare). \autoref{tab:dataset_examples} illustrates five example records with their domain and subject annotations.

\subsection{Statistical Analysis of Subject Annotations}

Sourced from an actual public library, this dataset reflects real-world practice and long-term quality constraints, shaped by evolving staff and workflows.
Nevertheless, it offers a valuable resource for building reliable AI tools for librarians—tools designed around corpus-level patterns and principled reasoning rather than idiosyncratic particularities. Thus, in this section, we statistically characterize the subject space—quantifying split overlap and long-tail sparsity, measuring distributional divergence (KL, JSD, and $\chi^2$), and assessing polysemy—to surface implications for model design and evaluation.

\subsubsection{Overlap and Long-Tail Phenomenon}
\label{sec:longtail}

From the nearly 200{,}000 subjects in the GND \textit{Sachbegriff}, 41{,}218 unique subjects appear in our dataset (per-split counts released \href{https://github.com/sciknoworg/tib-sid/blob/main/library-records-dataset/data-analysis/analysis1/subjects_by_split.csv}{here}). The annotations span traditional subjects (\textit{Literatur}, \textit{Architektur}, \textit{Philosophie}) and contemporary ones (\textit{Digitalisierung}, \textit{Nachhaltigkeit}, \textit{Künstliche Intelligenz}); 6{,}164 subjects occur in \emph{all three} splits, evidencing pronounced long-tail sparsity in which high-frequency labels (e.g., \textit{Literatur}, \textit{Architektur}, \textit{Unternehmen}) coexist with specialized ones (e.g., \textit{Robotik}, \textit{Bioinformatik}, \textit{Feminismus}), requiring models to handle few-shot and zero-shot generalization rather than rely on balanced head classes. Overall, the corpus bridges humanities, social sciences, and technical domains but is dominated by a long tail that should inform training and evaluation design. 
The Jaccard \citeyearpar{jaccard1912distribution} and weighted Jaccard (Tanimoto \citeyear{tanimoto1958elementary}) in \autoref{tab:split_overlap_summary} quantify split overlap: Jaccard gives the fraction of shared subjects (here $\approx 0.36$, i.e., about one third), while the weighted variant also accounts for subject frequencies. Lower Tanimoto reflect that even shared subjects occur at different rates, implying models must handle sparse labels and distribution shift.

\begin{table}[!htb]
\centering
\footnotesize
\begin{tabular*}{\columnwidth}{@{\extracolsep{\fill}}lccc@{}}
\hline
\textbf{Metric} & \textbf{Train--Dev} & \textbf{Train--Test} & \textbf{Dev--Test} \\
\hline
Jaccard & 0.3687 & 0.3608 & 0.3668 \\
Tanimoto & 0.2064 & 0.2530 & 0.3771 \\
\hline
\end{tabular*}
\caption{Jaccard and Tanimoto scores showing subject overlap across dataset splits.}
\label{tab:split_overlap_summary}
\end{table}

\subsubsection{Distributional Divergence}
\label{sec:distdivergence}

To move beyond overlap, we quantify frequency divergence across splits with KL divergence \cite{kullback1951information}, Jensen–Shannon divergence (JSD; \href{https://en.wikipedia.org/wiki/Jensen%E2%80%93Shannon_divergence}{link}), and the Chi-Squared ($\chi^2$) test \cite{pearson1900x}. KL spans 2–5 nats—$\mathrm{KL}(\text{Train}|\text{Dev}){=}4.34$, $\mathrm{KL}(\text{Train}|\text{Test}){=}4.69$ vs. $\mathrm{KL}(\text{Dev}|\text{Train}){=}1.99$, $\mathrm{KL}(\text{Test}|\text{Train}){=}1.89$—indicating Train covers a broader subject set while Dev/Test are narrower reweightings. 
JSD indicates moderate shifts (0.16 Train–Dev, 0.18 Train–Test, 0.26 Dev–Test). 
$\chi^2$ statistics of $\approx 37\text{k}$–$49\text{k}$ with $p{<}0.001$ confirm differences are systematic rather than random variation. Collectively, the corpus exhibits partial overlap but clear distributional drift. Methodologically, models must handle uneven frequencies and underrepresented topics—favoring regularization, calibration, and shift-aware validation and reporting.

\subsubsection{Assessing Polysemy}

To assess potential polysemy, we operationalize it as string identity shared across distinct GND identifiers and scan the taxonomy accordingly. Restricting to \emph{preferred labels}, such cases are exceedingly rare—65 occurrences (0.03\%) among 207{,}001 codes; for example, \textit{Alakaluf} is the preferred label for both \texttt{gnd:1071000497} (ethnographic group, Kawésqar) and \texttt{gnd:4001008-9} (broader folk-cultural classification), indicating parallel cataloging rather than genuine sense ambiguity. Including \emph{alternate labels} raises the incidence to 2{,}181 (0.52\%): \textit{Abwasseranalyse,Kennzahl} (“wastewater analysis, indicator”) spans seven entries—\texttt{gnd:4145594-0} (\textit{Biochemischer Sauerstoffbedarf}), \texttt{gnd:4147637-2} (\textit{Chemischer Sauerstoffbedarf}), \texttt{gnd:4185748-3} (\textit{Totaler Sauerstoffbedarf}), \texttt{gnd:4360165-0} (\textit{DOC}), \texttt{gnd:4381803-1} (\textit{TOC}), \texttt{gnd:4586442-1} (\textit{Pges.\ ICP}), \texttt{gnd:4586443-3} (\textit{TNb})—each denoting a measurement indicator; similarly, \textit{SPSS,WINDOWS,Programm} appears across five records for successive software versions. These findings suggest that duplicate strings predominantly reflect terminological reuse across related entities rather than true polysemy; consequently, alternate labels broaden retrieval but should not be treated as interchangeable with preferred labels in downstream modeling and evaluation.

We now go beyond exact name matches and use semantic embeddings to surface near-duplicates and potential sense conflations. Before encoding, we exclude acronym/code-like labels (fragile under distributional similarity), leaving $N=203{,}763$ subjects from $207{,}001$. We compare three embedding families (\autoref{tab:polysemy_summary}): Google’s \emph{EmbeddingGemma-300m} (compact multilingual, general-purpose representations) \cite{vera2025embeddinggemma}, \emph{multilingual-E5-small} (contrastive, retrieval-oriented) \cite{wang2024multilingual}, and \emph{jina-v2-base-de} (German-focused, high-precision bilingual encoder) \cite{mohr2024multi-jinai}. In all cases we construct a graph where an undirected edge links two subjects whose cosine similarity exceeds $0.90$; degree $\ge 2$, i.e. if a subject node has more than 2 similar terms, is a more pronounced signal of possible polysemy than isolated pairs.

\begin{table}[!htb]
\centering
\setlength{\tabcolsep}{1.5pt}      
\renewcommand{\arraystretch}{0.96} 
\begin{tabular}{@{}P{1.4cm}C{1.05cm}C{1.35cm}C{1.25cm}C{1.15cm}C{1.0cm}@{}}
\hline
\textbf{Model} & \textbf{View} & \textbf{N($\geq$1)} & \textbf{N($\geq$2)} & \textbf{Mean} & \textbf{Max} \\
\hline
\multirow{2}{1.15cm}{\raggedright \href{https://huggingface.co/google/embeddinggemma-300m}{Gemma-300m}}
  & name    & $\sim$19.8K  & $\sim$4.0K   & 0.15  & 83 \\
  & context & $\sim$15.7K  & $\sim$3.6K   & 0.13  & 75 \\ \hline
\multirow{2}{1.15cm}{\raggedright \href{https://huggingface.co/intfloat/multilingual-e5-small}{E5-small}}
  & name    & $\sim$195.7K & $\sim$188.1K & 38.77 & 1.49K \\
  & context & $\sim$165.8K & $\sim$147.0K & 24.62 & 800 \\ \hline
\multirow{2}{1.15cm}{\raggedright \href{https://huggingface.co/jinaai/jina-embeddings-v2-base-de}{Jina-v2-DE}}
  & name    & $\sim$15.0K  & $\sim$2.8K   & 0.11  & 36 \\
  & context & $\sim$14.5K  & $\sim$3.7K   & 0.13  & 53 \\
\hline
\end{tabular}
\caption{Semantic-similarity graph at cosine $\geq 0.90$ (topK$=50$), $N=203{,}763$ subjects.
N($\geq$1): count of subjects that have at least one similar neighbor; N($\geq$2): subjects with at least two similar neighbors (stronger polysemy signal).}
\label{tab:polysemy_summary}
\end{table}

Results in \autoref{tab:polysemy_summary} show clear differences across encoders. E5 produces very dense neighborhoods (name view: $\sim195.7\text{K}$ nodes with neighbors; mean degree $38.77$; max $1.49\text{K}$), which is useful for retrieval but tends to overstate polysemy by grouping topical associates and word-family variants rather than only near-duplicates. This is visible in concrete cases: \emph{Vergrößerung} links to $76$ items mixing near-synonyms (\emph{Erweiterung, Erhöhung}), antonyms (\emph{Verkleinerung}), derivational relatives (\emph{Verfilmung, Versilberung}), and broad associates (\emph{Darstellung, Durchmesser}); \emph{Zeit} pulls a long tail of compounds/contexts (\emph{Zeitmessung, Zeitraum, Zeitalter, Zeitzone, Zeitarbeit, \dots}); \emph{Differenzierung} balloons to $160$ neighbors across mathematics, sociology, and linguistics. By contrast, Gemma and Jina yield much sparser graphs (mean degree $\approx 0.11$–$0.15$, max $\le 83$), aligning better with our goal of flagging only very similar terms. Their links are tight: Gemma connects \emph{Zukunft}~$\to$~\emph{Futur}, \emph{Nutzung}~$\to$~\emph{Benutzung}/\emph{Nutzungseignung}, \emph{Instrument}~$\to$~\emph{(Musik)instrument}, with occasional orthographic/morphemic effects (\emph{Vergrößerung}~$\to$~\emph{Vergröberung}; \emph{Zeit}~$\to$~\emph{Zeitmaß}/\emph{Zeitmittel}/\emph{Saatzeit}). Jina is stricter still at $0.90$, mostly surfacing crisp pairs (\emph{Nutzung}~$\to$~\emph{Benutzung}; \emph{Differenzierung}~$\to$~\emph{Dedifferenzierung}; \emph{Instrument}~$\to$~\emph{Musikinstrument}/\emph{A-Instrument}) while many high-level nouns (\emph{Verhandlung, Entwicklung, Werk, Ziel}) have degree $0$—behaviour that is desirable if the aim is to flag only plausible sense conflations. For the rows against the view \emph{context} in \autoref{tab:polysemy_summary}, we embed a compact string that joins the preferred label with short definitional cues and a capped list of alternates (\texttt{Name [SEP] DEF:\dots [SEP] ALT:\dots}). For Gemma and E5 this generally sharpens meanings and prunes superficial, name-only links, whereas for Jina it sometimes consolidates genuine German paraphrases into small, tighter clusters—exactly the kind of behavior desired for a high-precision polysemy screen. Overall, polysemy in the GND appears rare: with conservative encoders (Gemma/Jina) at a strict $0.90$ cosine, only $\approx$1–2\% of subjects participate in multi-item clusters (degree $\ge 2$) while about 90–93\% have no near neighbor at all—indicating that genuine ambiguity is exceptional rather than pervasive.

\section{Three Systems}
\label{sec:systems}

To illustrate how the dataset can be used for automated subject indexing, we report results from three representative systems evaluated in the LLMs4Subjects shared tasks at \href{https://sites.google.com/view/llms4subjects}{SemEval~2025} \cite{dsouza-etal-2025-semeval} and \href{https://sites.google.com/view/llms4subjects-germeval/home}{GermEval~2025}, both of which used this dataset. The task is formulated as follows: given a record consisting of a title and abstract, a system should retrieve up to 20 relevant subjects from the GND taxonomy, ranked in descending order of confidence. This constitutes an information retrieval (IR) problem with ranked outputs, evaluated over the top-$k$ predicted subjects \cite{schutze2008introduction}. As the primary evaluation metric, we use nDCG@k (Normalized Discounted Cumulative Gain) \cite{jarvelin2002cumulated}, a rank-based measure that assesses how closely a system's predicted subject ranking matches the ground truth. Unlike recall@k, which measures only how many relevant subjects appear among the top-$k$ predictions, nDCG@k additionally considers their \textit{positions} in the ranking, rewarding correct subjects that appear higher in the list through a logarithmic discount. The score is normalized between 0 and 1, where 1 is a perfect ranking with all relevant subjects at the top.

In practice, this task is commonly addressed using XMTC methods. XMTC methods for controlled-vocabulary subject indexing typically combine scalable candidate generation over very large label spaces with a ranking stage optimized for top-$k$ predictions. Early approaches relied on sparse one-vs-rest linear classifiers such as DiSMEC \cite{babbar2017dismec} and PD-Sparse/PPD-Sparse \cite{yen2016pdsparse,yen2017ppdsparse}, which train independent linear models for each label while exploiting sparsity and distributed optimization to scale to very large label sets. Tree-based methods such as FastXML \cite{prabhu2014fastxml}, Parabel \cite{prabhu2018parabel}, and Bonsai \cite{khandagale2020bonsai} improved efficiency by organizing labels into hierarchical partitions, allowing inference to focus on a much smaller candidate subset at prediction time. Embedding and nearest-neighbor approaches such as SLEEC \cite{bhatia2015sparse} and AnnexML \cite{tagami2017annexml} instead learn low-dimensional representations for instances and labels, reframing XMTC as a semantic retrieval problem in which relevant labels are recovered via nearest-neighbor search. More recent neural methods, including XML-CNN \cite{liu2017deep}, AttentionXML \cite{you2019attentionxml}, X-Transformer \cite{chang2020taming}, and XR-Transformer \cite{zhang2021fast}, combine deep text encoders with label shortlisting to improve contextual semantic matching while preserving scalability. For library subject indexing, these families of methods can be understood as variants of a common retrieve-and-rank design pattern: first narrowing the controlled vocabulary to plausible candidate subjects, then estimating which labels are most appropriate for final assignment. Toolkits such as PECOS \cite{yu2022pecos} make this connection explicit by unifying hierarchical indexing, learned matching, and ranking within modular \textit{index $\rightarrow$ match $\rightarrow$ rank} pipelines.

Given this task formulation and methodological landscape, we focus on three representative systems from the LLMs4Subjects shared tasks: LA$^2$I$^2$F~\cite{salfinger-etal-2025-la2i2f}, KIFSPrompt~\cite{kahler-etal-2025-dnb}, and Annif~\cite{suominen-etal-2025-annif-germeval}. Together, they capture key design patterns that emerged in the shared tasks, from prompting-based pipelines to hybrid trained XMTC systems. Beyond these three, the broader submissions covered a diverse methodological space, including retrieval-only pipelines that transfer subjects from similar indexed records \cite{tian2025ructeam,singh2025silp_nlp}, bi-encoder plus cross-encoder reranking pipelines that first retrieve candidates and then rescore them more precisely \cite{dorkin2025tartunlp}, multilingual BERT ensembles that combine predictions from several language-specific or multilingual encoders \cite{hahn2025jim}, contrastively finetuned embedding models that explicitly pull relevant record--subject pairs closer in representation space \cite{jiang2025ynu}, Burst Attention--enhanced retrieval methods that refine embeddings by modeling interactions across embedding dimensions before top-$k$ retrieval \cite{islam2025nbf}, RAG-style subject selection pipelines that retrieve candidate subjects and then use an LLM to verify or rank them in context, leveraging the OntoAligner Python toolkit~\cite{tekanlou2025homa,babaei2025ontoaligner}, and LLM finetuning pipelines with synthetic data generation and preference optimization to better align outputs with human indexing behavior \cite{tian2025dutir831,ho-2025-ubffm}. Across these submissions, strong performance was typically associated with multi-stage architectures rather than single-model prediction alone, with common strategies including candidate retrieval followed by reranking, model ensembling, multilingual processing, and LLM-assisted augmentation or refinement. The three systems presented below were selected not because they exhaust this design space, but because they provide clear, interpretable exemplars of its main families and thus serve well as reference points for understanding the dataset.

\subsection{Approaches}

\paragraph{System 1~\cite{salfinger-etal-2025-la2i2f}.}
Library corpora commonly exhibit long-tailed subject distributions: a few high-level subjects occur frequently, while many fine-grained subjects appear rarely. Models trained directly on such data tend to overpredict frequent subjects and overlook rare but meaningful ones. To mitigate this, system~1 reframes multi-label classification as semantic retrieval in a shared 768-dimensional embedding space using \href{https://huggingface.co/sentence-transformers/all-mpnet-base-v2}{sentence-transformers/all-mpnet-base-v2} \cite{song2020mpnet}. Both training records (title + abstract) and subject labels (names + alternates) are embedded into the same space to enable semantic comparison with a query record. The method combines two complementary strategies: \textit{\underline{ontological reasoning}}, which compares the query embedding directly to subject embeddings, and \textit{\underline{analogical reasoning}}, which retrieves similar training records and transfers their assigned subjects. Candidate subjects from both sources are merged, deduplicated, and ranked by embedding distance, with the top results predicted.

System~2 builds upon system~1’s analogical reasoning thread with a few-shot retrieval strategy.

\paragraph{System 2~\cite{kahler-etal-2025-dnb}.}
This approach implements a four-stage pipeline combining retrieval-augmented few-shot prompting with controlled vocabulary mapping and ranking. 
\textit{\underline{Retrieve.}} The input record is embedded with the multilingual BGE-M3 model \cite{chen2024bge}, and a \href{https://weaviate.io/}{Weaviate vector store} retrieves the $L$ most similar training documents in the same language. 
\textit{\underline{Complete.}} These examples, together with their GND subject annotations, are inserted into an on-the-fly few-shot prompt for the \href{https://huggingface.co/mistralai/Ministral-8B-Instruct-2410}{Ministral-8B-Instruct model}, which generates free-form keyword suggestions. 
\textit{\underline{Map.}} The generated keywords are embedded with BGE-M3 and mapped to GND subjects using hybrid HNSW + BM25 search \cite{robertson2009bm25,malkov2018hnsw}, aligning free vocabulary to the controlled taxonomy. 
\textit{\underline{Rank.}} Finally, \href{https://huggingface.co/meta-llama/Llama-3.1-8B-Instruct}{Llama-3.1-8B-Instruct} \cite{dubey2024llama3} assigns each mapped term a 0–10 relevance score used to normalize and rank the final subject list. Like system~1, this pipeline relies entirely on off-the-shelf models without fine-tuning and performs only a single generation and ranking pass per record.

Finally, in contrast to systems~1 and~2, system~3 combines LLMs with traditional XMTC algorithms via the \href{https://annif.org/}{ANNIF toolkit} \cite{suominen2019annif}.

\paragraph{System 3~\cite{suominen-etal-2025-annif-germeval}.}
This hybrid approach uses LLMs for data preprocessing and final reranking while relying on XMTC models for subject prediction. 
\textit{\underline{Translate.}} Records and vocabularies are first translated into monolingual English and German collections using \href{https://huggingface.co/google/gemma-3-4b-it}{google/gemma-3-4b-it} and \href{https://huggingface.co/CohereLabs/aya-expanse-8b}{CohereLabs/aya-expanse-8b}, respectively.
\textit{\underline{Synthesize.}} Several LLMs then generate synthetic training data, producing four additional records per original entry in both languages. 
\textit{\underline{Train and predict.}} Two monolingual ensembles (English and German) are trained using three Annif backends: \href{https://github.com/tomtung/omikuji}{Omikuji} Bonsai \cite{khandagale2020bonsai} for partitioned label-tree classification, \href{https://github.com/NatLibFi/Annif/wiki/Backend%3A-MLLM}{MLLM} (Maui-like \cite{medelyan2009human} lexical matching) for text–term matching, and XTransformer, a transformer-based XMTC ranking model within the PECOS framework \cite{yu2022pecos}. Each backend is trained separately on monolingual datasets created via LLM translation and combined into language-specific ensembles. 
\textit{\underline{Merge and rerank.}} Finally, predictions are reranked using \href{https://huggingface.co/mistralai/Mistral-Small-3.1-24B-Instruct-2503}{mistralai/Mistral-Small-3.1-24B-Instruct-2503}, and bilingual outputs are merged to produce the final ranked list of subjects.

\begin{table}[!tb]
\centering
\footnotesize
\setlength{\tabcolsep}{4pt}
\renewcommand{\arraystretch}{0.9}
\begin{tabular}{@{}|l|c|c|c|c|@{}}
\hline
\textbf{Team Name} & \multicolumn{4}{c|}{\textbf{nDCG@k}} \\
\cline{2-5}
 & \textit{k=5} & \textit{k=10} & \textit{k=15} & \textit{k=20} \\
\hline
System 1 & 0.3639 & 0.3977 & 0.4143 & 0.4247 \\
System 2 & 0.4919 & 0.4880 & 0.4879 & 0.4879 \\
System 3 & 0.6020 & 0.6391 & 0.6560 & 0.6652 \\
\hline
\end{tabular}
\caption{nDCG@k scores at different ranked cutoffs for three systems on our subject indexing dataset.}
\label{tab:ndcg_overall}
\end{table}

\begin{figure}[!tb]
\begin{center}
\includegraphics[width=\columnwidth]{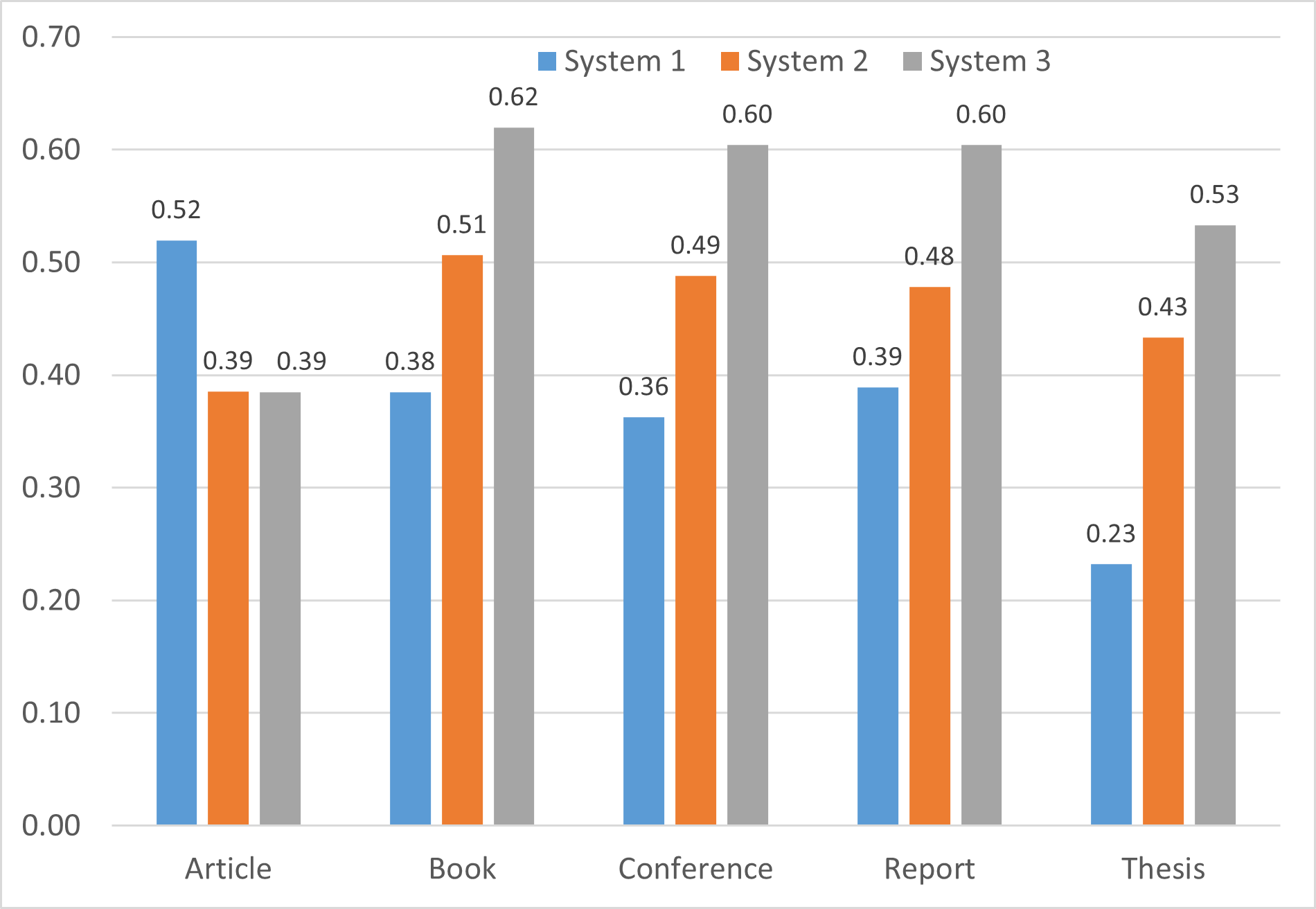}
\caption{nDCG@5 scores by the five record types.}
\label{fig:record-type}
\end{center}
\end{figure}

\begin{figure}[!tb]
\begin{center}
\includegraphics[width=0.65\columnwidth]{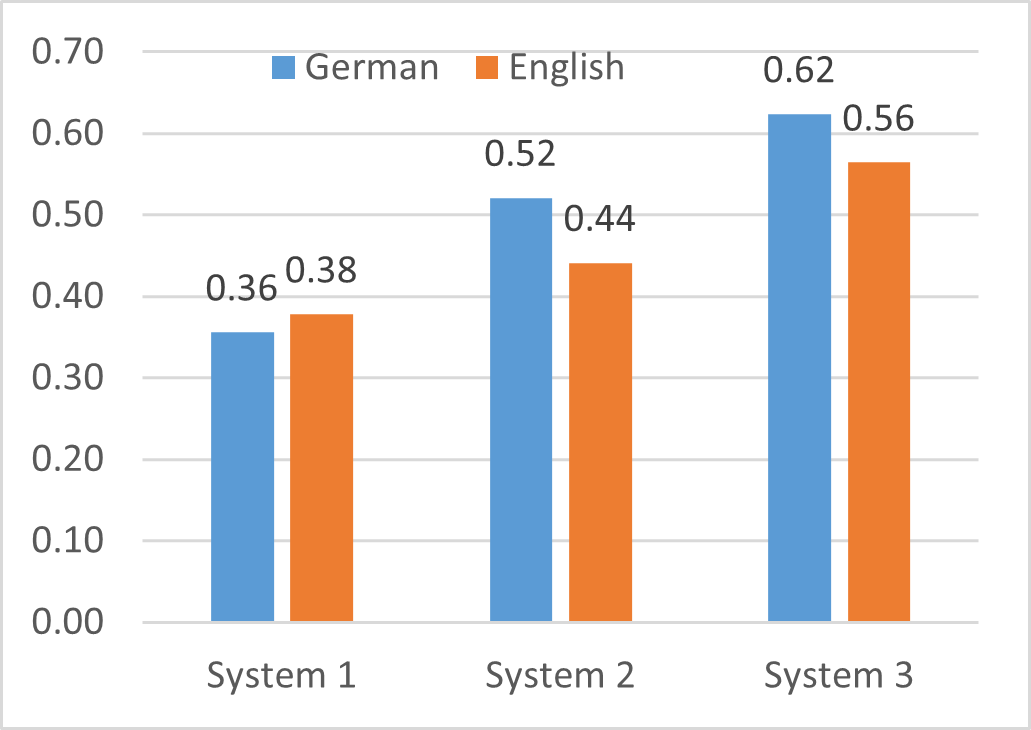}
\caption{nDCG@5 scores by the two languages.}
\label{fig:language}
\end{center}
\end{figure}

\begin{figure*}[!tb]
\centering
\captionsetup[subfigure]{justification=centering}
\begin{subfigure}{\textwidth}
  \centering
  \includegraphics[width=\textwidth]{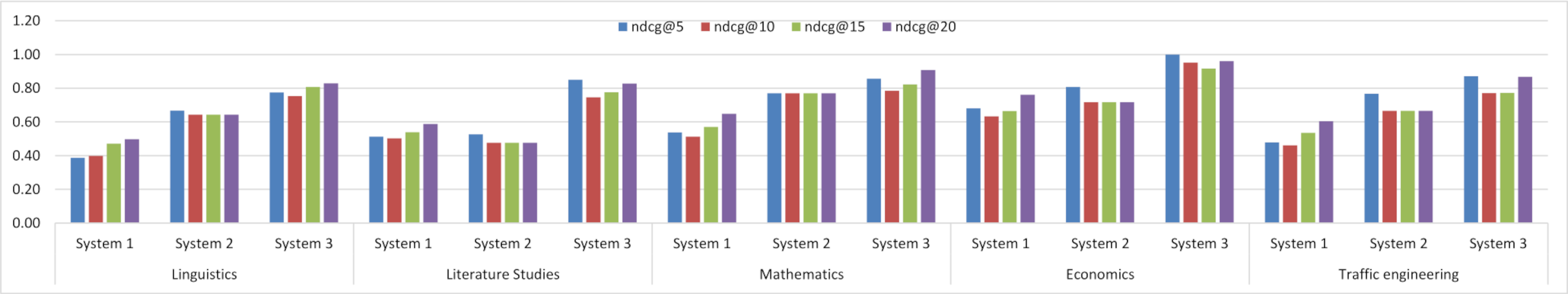}
  \caption{Case 1 - Treating both Y (correct) and I (irrelevant but technically correct) subject predictions as correct.}
  \label{fig:case:a}
\end{subfigure}
\vspace{0.6em}
\begin{subfigure}{\textwidth}
  \centering
  \includegraphics[width=\textwidth]{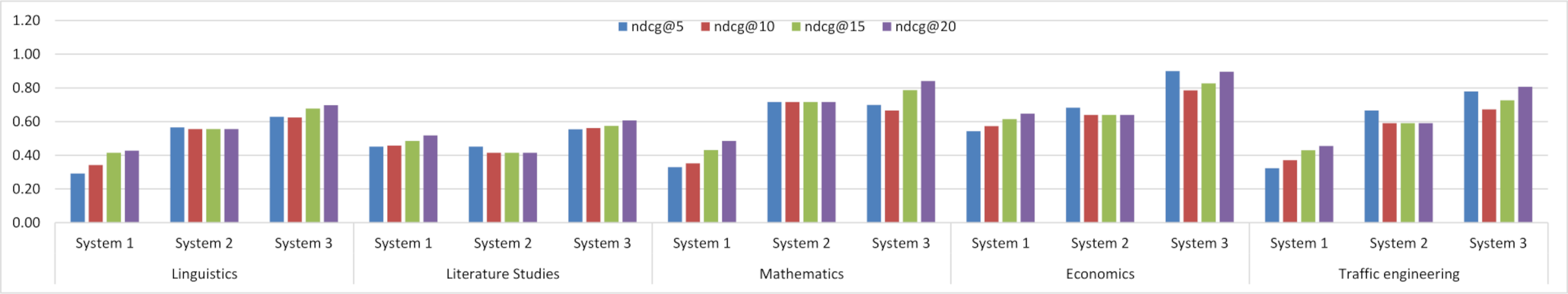}
  \caption{Case 2 - Treating only Y (correct) subject predictions as correct.}
  \label{fig:case:b}
\end{subfigure}
\caption{nDCG@k scores, where k=5,10,15,20, for qualitative evaluation on 10 records per five domains. For this exercise, subject predictions were manually labeled Y and I by subject specialists at the library.}
\label{fig:qualitative-results}
\end{figure*}

\subsection{Quantitative Results}

\autoref{tab:ndcg_overall} presents the quantitative results on our released test set for the three systems. Overall, the heavily engineered System 3 proved most effective, while System 2 showcased a clever extension of System 1’s ideas by using dynamically retrieved few-shot examples to elicit strong task performance from LLMs without training. In practical library settings, the usefulness of AI assistance depends on having the most relevant subjects appear early in a concise prediction list. Based on librarian feedback, we set the optimal list length to $k=20$, balancing coverage and efficiency. \autoref{fig:record-type} and \autoref{fig:language} show performance by record type and language at nDCG@5. System 1 performed best on article records, while System 3 led across the other four types. By language, System 1 was strongest on English records, whereas Systems 2 and 3 performed better on German. Overall, System 2 establishes a baseline for untrained LLM-based methods (nDCG@5 = 0.4919), and System 3 sets the benchmark for hybrid, model-trained approaches (nDCG@5 = 0.6020).

\subsection{Qualitative Results}
\label{sec:qualitativeresults}
We asked subject librarians to manually assess outputs from the three systems as a form of human spot-checking (\autoref{fig:qualitative-results}). Ten random test records across five domains—Linguistics, Literature Studies, Mathematics, Economics, and Traffic Engineering—were each reviewed by a specialist. For every predicted subject, librarians marked Y (correct), I (technically correct but irrelevant), or N (incorrect). \autoref{fig:case:a} shows nDCG scores at k=5,10,15,20 when Y and I are treated as correct (case 1), while \autoref{fig:case:b} shows results when only Y counts as correct (case 2).

For case 1 and 2, the relative domain-wise nDCG@20 rankings show consistent patterns across systems. System 1 performed best in \textit{Economics}, followed by \textit{Mathematics} and \textit{Literature Studies}, while \textit{Traffic Engineering} lagged behind. System 2 achieved its strongest results in \textit{Mathematics} and \textit{Economics}, showing generally balanced performance across domains. System 3 outperformed both, reaching near-perfect scores in \textit{Economics} and \textit{Mathematics} and maintaining strong results across all domains, reflecting the advantage of its hybrid learning and ensemble design. The differences between case 1 (Y + I counted) and case 2 (Y only) reveal that many high-ranked subjects were \textit{contextually related} but not exact matches—indicating that systems, especially LLM-based ones, capture topical proximity well but still struggle to distinguish semantically distinct subjects within a record. This highlights a key challenge for future models: promoting conceptual diversity among top-ranked predictions rather than clustering around near-synonyms.

A closer look at System~1's output for these spot checked records revealed that the analogical reasoning branch dominated due to its similarity computation: embeddings of training records yielded more similar distances to test records than embeddings of GND subject terms. Consequently, correctly identified subjects from the ontological branch were often ranked lower. Another common error occurred when the analogical branch transferred \textit{all} subjects from a similar training record, even though only some applied. For example, in \textit{``Sprachwissenschaft im Fokus''}, the system inferred \textit{Englisch}, \textit{Spanisch}, and \textit{Romanische Sprachen} merely because they co-occurred with the correct label \textit{Deutsch} in the retrieved record, leading to many false positives and underscoring the need for filtering contextually relevant subjects. For System~2, most errors arose in the \textit{Map} component linking LLM-generated keywords to GND terms, including mapping errors due to unknown or ambiguous concepts in the GND; and correctly mapped but irrelevant yet technically valid suggestions (category~I). Overall, the system tended to produce plausible but overly broad terms rather than precise subject matches, which was also the case in many of system 3 errors.

\section{Conclusion}
\label{sec:conclude}

This paper introduced an XMTC dataset of library records paired with a rich subject taxonomy, offering a novel resource for subject indexing research. We presented three complementary approaches.
The best scoring system relied heavily on traditional machine learning, raising questions about the applicability and generalizability of purely LLM-based approaches, which remain largely untapped. Our dataset provides a challenging benchmark for assessing how LMs capture the nuanced semantics required for library indexing. Future work will benchmark multilingual embeddings, explore small LMs as efficient alternatives, and investigate LM distillation for practical deployment. It also supports evaluating subject indexing by hierarchy, transparency, and usefulness in real library work.

\nocite{*}
\section{Bibliographical References}\label{sec:reference}

\bibliographystyle{lrec2026-natbib}
\bibliography{paper-references}

\appendix

\section{System 1 Detailed Error Analysis}

In the following section, we present a more detailed analysis of the different systematic types of errors made by System~1 (LA$^2$I$^2$F)~\cite{salfinger-etal-2025-la2i2f} on the test data set described in \autoref{sec:qualitativeresults}. In this qualitative analysis, subject-matter librarians manually rated the system's outputs by checking the 20 predicted labels for each of 50 test documents, covering 161 gold-standard subjects. Each prediction was marked as either Y (correct), I (technically correct but irrelevant), or N (incorrect), allowing us to analyze systematic error patterns in greater depth.

\paragraph{False Negatives (Missed Ground Truth Labels).} 
In total, System~1 achieved a \emph{False Negative Rate}~(FNR) of 56.5\%, i.e., the share of missed ground-truth subject labels, when considering $k=20$ predicted subjects per document, and an FNR of 73.2\% when considering only the system's top-5 ranked predictions. However, due to the sheer size of the GND ontology and the human-tagged training corpus, the specific selection of subject labels assigned to a document can be somewhat subjective and thus biased by the tagging expert, especially if synonymous concepts are present in the ontology. To account for such potential ambiguities when evaluating the subject labels missed by the system's predictions, we dissect our analysis of \emph{False Negatives}~(FNs), i.e., ground-truth labels missed by the system, according to the following categories:
\begin{description}
\small
  \item[FNM:] aspect completely missed by the system
  \item[FNC:] aspect missed, but closely matched by other predictions
\end{description}

By manually checking System~1's top-5 ranked predictions only, we obtain the following absolute counts on the 50 test set records using these definitions:

\begin{table}[!htb]
  \centering
  \scriptsize
  \begin{tabular}{lr}
  \hline
  \textbf{Error Category} & \textbf{Absolute Count} \\
  \hline
  FNM & 79 \\
  FNC & 41 \\
  \hline
  False Negatives (total) & 120 \\
  \hline
  \end{tabular}
  \caption{Distribution of false negatives for 250 subject suggestions from System~1 (5 predicted labels per test set document) on 50 test-set documents with 161 gold-standard subjects.}
  \label{tab:error_categories_sys1}
\end{table}

As this fine-grained manual inspection, factoring in synonymous subjects, reveals, 34.1\% of System~1's FNs did not retrieve the exact ground-truth label but instead a closely matching synonym, while 65.8\% correspond to missing ground-truth concepts that are completely absent from System~1's top-5 predictions. 

When digging further into the reasons behind this, we identify the following systematic error category introduced by System~1: In System~1's implemented fusion strategy, \emph{analogical reasoning dominates}. Analyzing the contributions from System~1's two reasoning branches to the fused results, we observe that the analogical branch drives the results: document-to-document distances tend to be smaller than document-to-subject distances. Subjects contributed from the ontological branch thus are typically ranked behind those from the analogical branch and therefore often do not make it into the fused list of top-$k$ ranked subjects returned, as illustrated by the test set record shown in \autoref{tab:Sys1ex885} (its gold-standard predictions can be found in \autoref{tab:gold-standard-example}). In this case, the ground-truth label \textit{AWACS} was predicted by the ontological branch at rank~22, but got outranked in the fusion process by the smaller document distances, i.e., higher cosine similarities, of the analogical branch, resulting in a correct prediction being filtered out in the merging process. This indicates the need for developing a more sophisticated fusion strategy, which could include another downstream LLM-based relevancy filtering and re-ranking component, similar to System~2.

\paragraph{False Positives (Incorrect Predictions).} 
An additional downstream relevancy filtering component will also be needed to address another inherent error category identified: System~1 currently implements an overly optimistic assumption for analogical reasoning, considering \emph{all} subject tags from the closest training documents in embedding space as appropriate tags for the new document. We found this to be the dominant inherent error source for introducing false predictions not matching the new document, i.e., \emph{False Positives}~(FPs), in System~1. While related documents might overlap in their subject labels, not necessarily \emph{all} subjects from document~A might be appropriate for a similar document~B. 

\paragraph{Error Categories.} In summary, we denote the dominant identified systematic error categories as:
\begin{description}
\small
  \item[FPAR] FP introduced by analogical reasoning
  \item[OROR] Correct result from ontological reasoning outranked after fusion with analogical results
\end{description}

\autoref{tab:Sys1ex885} illustrates some concrete examples of these systematic errors on test set record ID \texttt{3A1007389885}.

\begin{table}[!h]
  \centering
  \scriptsize
  \begin{tabular}{p{2cm}*{4}{c}>{\centering\arraybackslash}p{1.5cm}}
    \hline
    \textbf{Prediction} & \textbf{R} & \textbf{Rank} & \textbf{Sim.} & \textbf{Eval.} & \textbf{Error}\newline \textbf{Category}\\
    \hline
        Elektronische Gegenmaßnahme \newline [electronic countermeasure] \newline \texttt{gnd:1082384615} & A & 1 & 0.687 & N & FPAR\\
        St\"orsender \newline [jammer] \newline \texttt{gnd:4273774-6} & A & 2 & 0.687 & Y & -- (TP, Y) \\
        Radar \newline [radar] \newline \texttt{gnd:4176765-2} & A & 3 & 0.687 & Y & -- (Y) \\
        Sonar \newline [sonar] \newline \texttt{gnd:4181785-0} & A & 4 & 0.652 & N & FPAR\\
        Signalverarbeitung \newline [signal processing] \newline \texttt{gnd:4054947-1} & A & 5 & 0.652 & Y & -- (Y) \\
    \hline
    AWACS \newline [airborne warning and control system] \newline \texttt{gnd:4309079-5} & O & 22 & 0.406 & & OROR\\
    \hline
  \end{tabular}
  \caption{System~1's predictions for the test set record with ID \texttt{3A1007389885}, \textit{``Recent advancements in airborne radar signal processing: emerging research and opportunities''}, dissecting which reasoning branch~(\textbf{R}) identified each predicted label -- analogical~(A) or ontological~(O) reasoning. The first five results are the top-5 predictions returned by the fusion system after merging the results from both A and O branches; the last row shows an actual ground-truth subject excluded from the retrieved list due to being outranked. \textbf{Rank} denotes the ranking of each subject within each reasoning branch (analogical vs.\ ontological), \textbf{Sim.} shows the cosine similarity to the query document in embedding space determining the ranking, and \textbf{Eval.} lists the human expert's judgment of the predicted label.}
  \label{tab:Sys1ex885}
\end{table}

In conclusion, this analysis confirms the hypothesized complementarity of System~1's fusion architecture, with both reasoning branches identifying complementary information, and identifies avenues for future work. The systematic error sources identified can be tackled by introducing additional filtering components for evaluating and re-ranking the identified (candidate) subjects, which should up-rank matching predictions contributed from the ontological branch and eliminate FPs introduced by the analogical branch.

\section{System 2 Detailed Error Analysis}

Manual inspection of the system output reveals certain repeatedly occurring error patterns. In a close examination of the 50 test documents that were also rated by the subject experts during qualitative evaluation, we observe 176 subject terms suggested by System~2 (KIFSPrompt)~\cite{kahler-etal-2025-dnb} in total, including 110 false positives that can be classified into a variety of subcategories. The entire data sheet for the qualitative error analysis can be found at \url{https://github.com/sciknoworg/tib-sid/tree/main/evaluation/results/system2%20-%20additional%20analysis}.

Forty of the false positive suggestions were rated by the subject experts as relevant, and 27 suggestions were rated as not relevant but technically correct. This leaves 43 false positive suggestions (39\%) that should be considered truly erroneous. It is interesting to study how the system came to make these truly erroneous suggestions. Indeed, some of the errors can be attributed to issues occurring during the mapping stage of the system, where free keywords get matched to normalized subject terms. We observe two recurring patterns:

\begin{description}
\small
  \item[MEU:] mapping error, because a concept is unknown in the GND
  \item[MEA:] mapping error, because a concept is ambiguous in the GND
\end{description}

Let us illustrate these categories with examples. In our sample, a document titled \textit{"The Arden research handbook of Shakespeare and adaptation"} is tagged with the gold-standard subject term \textit{Adaption (Literatur)} (gnd:102289935X). The LLM suggested simply \textit{Adaption}, which is used as an alternative label for the GND subject term \textit{Anpassung} (gnd:4128128-7). The mapping used this alternative label and found a perfect match. This is what we mean by \textbf{MEA}: a mapping error due to ambiguity.

For an illustration of \textbf{MEU}, consider the title
\textit{"Recent advancements in airborne radar signal processing : emerging research and opportunities"} (record-ID 3A1007389885). \autoref{tab:sys2-example} shows the output of System~2.

\begin{table}[!htb]
  \centering
  \scriptsize
  \begin{tabular}{p{2.3cm}p{2.4cm}l}
    \hline
    \textbf{LLM-suggested term} & \textbf{Mapped GND term} & \textbf{Error Category}  \\
    \hline
        \textit{Signalverarbeitung} \newline [Signal processing] & Signalverarbeitung \texttt{gnd:4054947-1} \newline [Signal processing] & FP, but relevant \\
        Jamming  \newline [Jamming] & St\"orsender \texttt{gnd:4273774-6} \newline [jamming transmitter] & TP \\
        Spoofing \newline [Spoofing] & betr\"ugen \texttt{gnd:4554780-4} \newline [deceive] & MEU \\
        Luftfahrtradar  \newline [aviation radar] & Wetterradar \texttt{gnd:4270420-0} \newline [Weather radar] & MEU \\
    \hline
  \end{tabular}
  \caption{System~2 output for the record-ID \texttt{3A1007389885} \textit{``Recent advancements in airborne radar signal processing : emerging research and opportunities''} with error category annotation.}
  \label{tab:sys2-example}
\end{table}

\autoref{tab:gold-standard-example} shows the gold-standard annotations for the same document.

\begin{table}[!htb]
  \centering
  \scriptsize
  \begin{tabular}{p{2.3cm}l}
    \hline
    \textbf{Gold-standard term} & \textbf{GND identifier} \\
    \hline
    St\"orsender \newline [jamming transmitter] & \texttt{gnd:4273774-6} \\\hline
    AWACS \newline [airborne warning and control system] & \texttt{gnd:4309079-5} \\\hline
    Raum-Zeit-Signalverarbeitung \newline [space-time signal processing] & \texttt{gnd:4834654-8} \\\hline
    Bordradar \newline [on board radar] & \texttt{gnd:4456131-3} \\\hline
    Zielerkennung \newline [target recognition] & \texttt{gnd:4190792-9} \\\hline
    T\"auschung (Milit\"ar) \newline [deception (military)] & \texttt{gnd:4184333-2} \\
    \hline
  \end{tabular}
  \caption{Gold-standard terms for the record-ID \texttt{3A1007389885}.}
  \label{tab:gold-standard-example}
\end{table}

We see that the LLM-suggested candidate \textit{Spoofing} is a close match to the concept \textit{T\"auschung (Milit\"ar)}, and \textit{Luftfahrtradar} is close to both \textit{AWACS} and \textit{Bordradar}. However, as \textit{Spoofing} and \textit{Luftfahrtradar} do not exist directly in the GND, these terms get matched to the wrong entities in the GND. In these cases, the LLM suggested concepts that are unknown in the GND.

Out of the 43 truly erroneous suggestions in our sample, 11 may be classified as \textbf{MEU} and 5 may be classified as \textbf{MEA}.

Fine-tuning of the ranking stage and stricter filtering based on cosine similarity might help alleviate such errors in the future. Suggestions removed in such a filtering step could also be used as candidates for new subject terms in the GND that need to be added to the vocabulary (as synonyms or new concepts). The problem of resolving ambiguity remains a severe challenge. In a productive setting, analyzing these errors to enhance the vocabulary for automated subject indexing would complement the system for improved performance.

Complementing the analysis of false positives, we can also analyze this sample with a focus on recall and take a closer look at false negatives.

Using the false-negative categories introduced earlier\footnote{Opposed to System~1 and System~3, System~2 has a dynamically regulated number of suggestions per record, typically fewer than five. It is therefore irrelevant to analyze whether an aspect was matched later in the ranking, as for the other systems}, we find the following absolute counts:

\begin{table}[!htb]
  \centering
  \scriptsize
  \begin{tabular}{lc}
  \hline
  \textbf{Error Category} & \textbf{Absolute Count} \\
  \hline
  FNM & 68\\
  FNC & 28\\
  \hline
  False Negatives (total) & 96 \\
  \hline
  \end{tabular}
  \caption{Distribution of false negatives for 176 subject suggestions from System~2 on 50 test-set documents with 162 gold-standard subjects.}
  \label{tab:error_categories_sys2}
\end{table}

If we only count \textbf{FNM} as truly missing aspects, we find a micro average recall of
\[
\text{Rec} = \frac{\text{TP}}{\text{TP} + \text{FNM}} = 0.49
\]

\section{System 3 Detailed Error Analysis}

In a close examination of the 50 test-set documents that were also rated by the experts during qualitative evaluation and their top-5 subjects predicted by System~3 (Annif)~\cite{suominen-etal-2025-annif-germeval} (250 predicted subject terms in total), as well as their TIBKAT ground-truth subjects (161 in total), we found 91 true positives and 70 false negatives. In addition to the false-negative categories introduced earlier, we define the following additional category:

\begin{description}
\small
  \item[FNM5:] aspect missed by the top-5 predictions but present in the subsequent top-20 predictions (subcategory of FNM)
\end{description}

Using this classification, we found 61 cases of FNM, of which 25 were FNM5; that is, the subject did not appear among the top-5 predictions, but was present further down the list of the top-20 predictions. We also found 9 cases of FNC. The full data for the error analysis can be found at \url{https://github.com/sciknoworg/tib-sid/tree/main/evaluation/results/system3%20-%20additional%20analysis}.

To illustrate these categories with examples, consider the document titled \textit{"Dynamic term structure modeling beyond the paradigm of absolute continuity"} (record-ID 3A168734406X). \autoref{tab:sys3-example} shows the top-5 predictions of System~3 for this document.  
\autoref{tab:sys3-gold-standard-example} shows the gold-standard annotations for the same document.

\begin{table}[htb]
  \centering
  \scriptsize
  \begin{tabular}{p{1.9cm}lcp{1.65cm}}
    \hline
    \textbf{Prediction} & \textbf{GND identifier} & \textbf{Eval.} & \textbf{Error Cat.} \\
    \hline

        \textit{Kreditrisiko} \newline [credit risk] & \texttt{gnd:4114309-7} & Y & FP, but relevant \\
        \textit{Kreditmarkt} \newline [credit market] & \texttt{gnd:4073788-3} & Y & FP, but relevant \\
        \textit{Zinsstrukturtheorie} \newline [term structure theory] & \texttt{gnd:4117720-4} & Y & FP, but relevant \\
        \textit{Modellierung} \newline [modelling] & \texttt{gnd:4170297-9} & I & FP, technically correct \\
        \textit{Semimartingal} \newline [semimartingale] & \texttt{gnd:4180967-1} & Y & TP \\
    \hline
  \end{tabular}
  \caption{System~3 top-5 output for the record-ID \texttt{3A168734406X} \textit{``Dynamic term structure modeling beyond the paradigm of absolute continuity''} with expert evaluations and error category annotation.}
  \label{tab:sys3-example}
\end{table}

\begin{table}[htb]
  \centering
  \scriptsize
  \begin{tabular}{p{2.0cm}lcl}
    \hline
    \textbf{Gold-standard \newline subject} & \textbf{GND identifier} & \textbf{Freq.} & \textbf{Error Cat.} \\
    \hline

    Arbitrage \newline [arbitrage] & \texttt{gnd:4002820-3} & 0 & FNM \\
    Semimartingal \newline [semimartingale] & \texttt{gnd:4180967-1} & 4 & TP \\
    Ausfallrisiko \newline [default risk] & \texttt{gnd:4205942-2} & 4 & FNM5 \\
    Zinsstruktur \newline [term structure] & \texttt{gnd:4067855-6} & 1 & FNC \\
    HJM-Modell \newline [HJM model] & \texttt{gnd:4642940-2} & 1 & FNM5 \\

    \hline
  \end{tabular}
  \caption{Gold-standard subjects for the record-ID \texttt{3A168734406X} with train-set frequencies and error category annotation.}
  \label{tab:sys3-gold-standard-example}
\end{table}

This example document illustrates the difficulties of accurately predicting the gold-standard subjects, despite the predictions being rated as very relevant by the subject experts. Out of the top-five predictions in \autoref{tab:sys3-example}, only \textit{Semimartingal} appears in the gold-standard subjects and thus counts as a true positive. Of the other four predictions, three were considered relevant (Y) and one was considered technically correct but irrelevant (I) by the subject experts.

Looking at the same document from the perspective of the five gold-standard subjects in \autoref{tab:sys3-gold-standard-example}, again only one is a true positive while the other four are different kinds of false negatives: \textit{Arbitrage} was not predicted at all by the system, \textit{Ausfallrisiko} and \textit{HJM-Modell} were not in the top-five predictions but appeared within the remaining top-20, and \textit{Zinsstruktur} was not predicted, but the closely related concept \textit{Zinsstrukturtheorie} was among the top-five predictions. All five gold-standard subjects for this document have a low frequency in the training set, ranging from 0 to 4 occurrences. This makes it challenging to predict them using System~3, which mostly relies on models that learn to distinguish each individual subject based on patterns in the training data.

\begin{figure}[htb]                                                        
  \centering                                                             
  \includegraphics[width=\columnwidth]{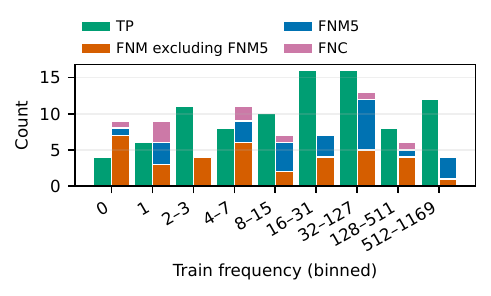}  
  \caption{Train-frequency distribution (binned) for the System~3 predictions, split into true positives and three false-negative subtypes.}        
  \label{fig:system3_train_freq_hist}                                    
\end{figure}

\autoref{fig:system3_train_freq_hist} shows a histogram of the subjects binned by training-set frequency and split into true positives and the three subtypes of false negatives (in the diagram, FNM5 has been separated out from the remaining FNM cases). For the low-frequency bins 0, 1, and 4--7, false negatives dominate over true positives. In the higher frequency bins from 8--15 upwards, true positives are more common than false negatives. Bin 2--3 is an outlier in this pattern where true positives dominate despite the low training-set frequency.

Despite the outlier bin, the pattern is clear: System~3 struggles in the prediction of subjects that have a low frequency in the training set, while higher-frequency subjects are more often correctly predicted. This is expected, because out of the three algorithms that form the ensemble, only one (MLLM) is capable of predicting zero-shot subjects, while the other two algorithms (Omikuji and XTransformer) rely on subject-specific training data and therefore cannot predict low-frequency subjects very well, or indeed at all in the zero-shot case. In the future, the system could be improved by including more methods focused on predicting low-frequency terms, for example by matching document text to GND subjects via embeddings as in System~1, or by using off-the-shelf LLMs to suggest possible candidate subjects as in System~2.
\end{document}